\documentclass{article}

\usepackage[preprint]{neurips_2025}

\usepackage{multirow}
\usepackage[utf8]{inputenc} 
\usepackage[T1]{fontenc}    
\usepackage{hyperref}       
\usepackage{url}    
\usepackage{graphicx}
\usepackage{booktabs}       
\usepackage{amsfonts}       
\usepackage{amsmath}        
\usepackage{nicefrac}       
\usepackage{microtype}      
\usepackage{xcolor}         
\usepackage{tcolorbox}
\usepackage{float}

\title{Training-Free Stylized Abstraction}

\author{%
  Aimon Rahman$^*$, Kartik Narayan$^*$, Vishal M. Patel \\
  Johns Hopkins University\\
  \texttt{\{arahma30, knaraya4, vpatel36\}@jhu.edu} \\
  { \textcolor{magenta}{\url{https://kartik-3004.github.io/TF-SA/}}} \\
}

\begin{document}

\maketitle

\renewcommand{\thefootnote}{\fnsymbol{footnote}}
\footnotetext[1]{Equal contribution}
\renewcommand{\thefootnote}{\arabic{footnote}}

\begin{figure*}[!htb]
\vskip -15pt
    \centering
    \includegraphics[width=1\textwidth]{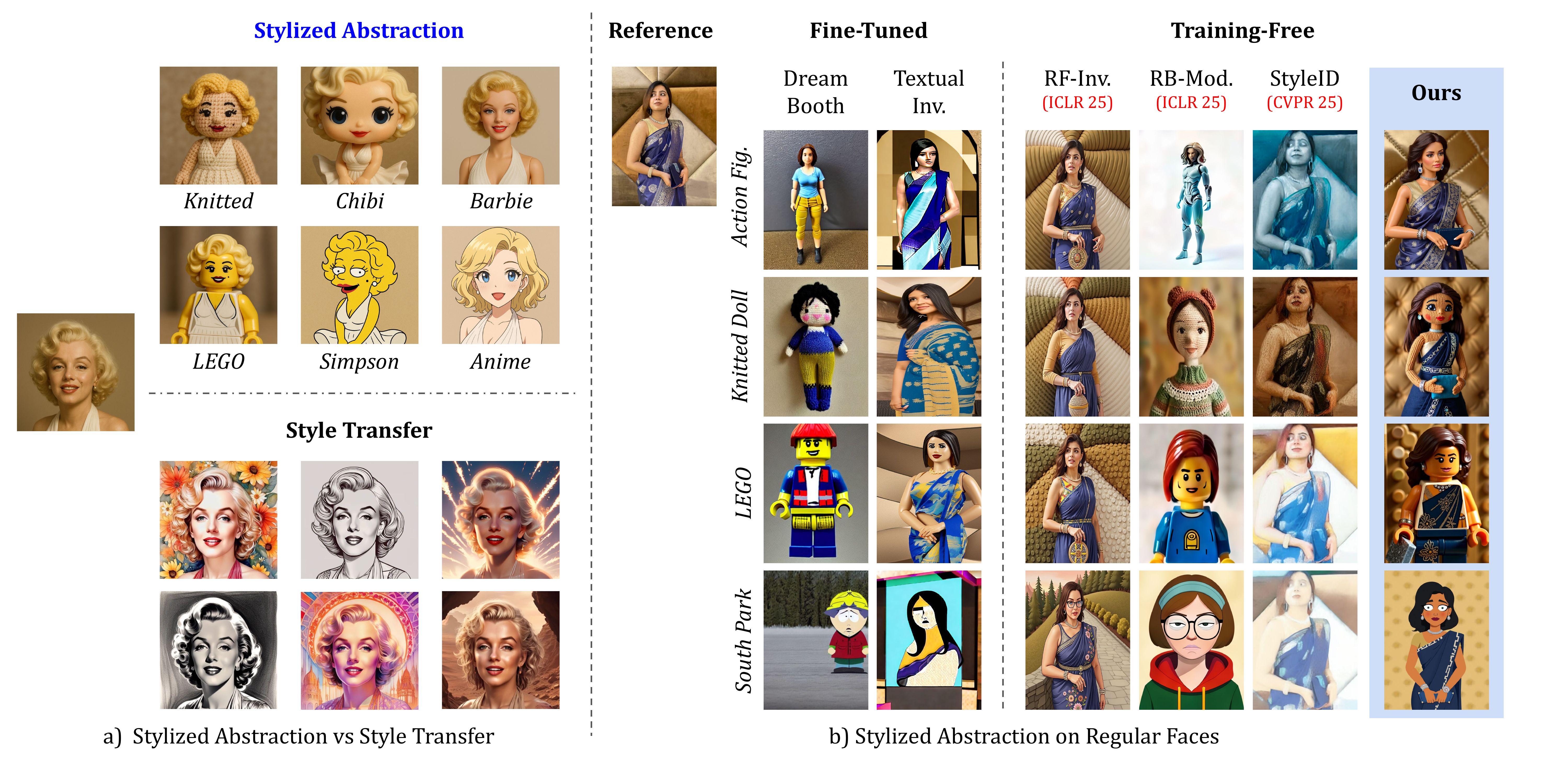}
    \vspace{-15pt}
    \caption { \textbf{a)} Style Abstraction vs. Traditional style transfer. \textbf{(Top)} Stylized abstraction techniques capture core identifying attributes while allowing stylistic distortion to preserve the intended visual style. \textbf{(Bottom)} Traditional style transfer preserves geometry and appearance but applies texture-based styles, often failing to generalize beyond appearance-level edits.
    \textbf{b)} Comparison across existing style transfer/personalized generation using a \textbf{single image} of a \textbf{non-celebrity subject}. Most methods struggle to retain semantic identity for everyday individuals, while our \textbf{training-free} method preserves key identity cues across diverse styles.}
    \label{fig:teaser}
\end{figure*}
\begin{abstract}
Stylized abstraction synthesizes visually exaggerated yet semantically faithful representations of subjects, balancing recognizability with perceptual distortion. Unlike image-to-image translation, which prioritizes structural fidelity, stylized abstraction demands selective retention of identity cues while embracing stylistic divergence, especially challenging for out-of-distribution individuals. We propose a training-free framework that generates stylized abstractions from a single image using inference-time scaling in vision-language models (VLLMs) to extract identity-relevant features, and a novel cross-domain rectified flow inversion strategy that reconstructs structure based on style-dependent priors. Our method adapts structural restoration dynamically through style-aware temporal scheduling, enabling high-fidelity reconstructions that honor both subject and style. It supports multi-round abstraction-aware generation without fine-tuning. To evaluate this task, we introduce \textit{StyleBench}, a GPT-based human-aligned metric suited for abstract styles where pixel-level similarity fails. Experiments across diverse abstraction (e.g., LEGO, knitted dolls, South Park) show strong generalization to unseen identities and styles in a fully open-source setup.
\end{abstract}

\section{Introduction}
\textbf{Image-to-image style translation} \cite{deng2022stytr2, sohn2023styledrop, wang2023stylediffusion, jiang2024diffartist, jing2019neural,xing2024csgo, chen2021artistic} is a well-studied area that traces its origins to GAN-based approaches, such as neural style transfer \cite{gatys2015neural} and CycleGAN \cite{zhu2017unpaired}, and has since evolved to include both diffusion-based training and training-free methods \cite{rout2025semantic, rout2025rbmodulation, le2022styleid, mo2024freecontrol, zhang2023adding, zhao2023uni,  deng2022stytr2, liu2023name, chen2024artadapter}. These techniques typically focus on overlaying a specific style onto an input image while preserving the subject's identity. Common examples include transforming portraits into sketches, cartoons, or artwork in the style of artists like Van Gogh. Importantly, the resulting stylized images often retain structural consistency with the original content.

\textbf{Stylized abstraction}, on the other hand, involves exaggerating or simplifying the features of a subject to create a stylized representation. Rather than aiming for photorealism, it emphasizes recognizable traits that evoke the subject’s \textbf{concept or identity} (Illustrated in Figure \ref{fig:teaser} (a)). Stylized representation aims to capture the essence of a subject through \textit{visual abstraction}, focusing less on exact likeness and more on the retention of key, recognizable features \cite{berger2013style}. For instance, a knitted doll or a LEGO figure of Einstein may omit intricate facial geometry or biometric precision, yet still be immediately identifiable due to consistent visual traits such as his distinctive hair, mustache, or attire. These features serve as semantic anchors, allowing viewers to recognize the subject even in highly abstracted or playful forms. This form of representation is widespread in media, animation, and merchandising, where retaining a character’s identity in a simplified, reproducible form is essential. Terms like \textit{personified toy representation} or \textit{iconic stylization} are often used to describe such instances. Unlike traditional image-to-image translation, which typically enforces structural consistency, stylized abstraction embraces simplification, distortion, or even exaggeration to evoke familiarity and conceptual identity.

Stylized abstraction, in contrast to traditional image-to-image translation, remains a relatively underexplored topic. The challenge lies in its nuanced nature. While image-to-image translation typically involves aligning an input image with the distribution of a target style, often while preserving geometric structure. This is a relatively simpler task. For example, sketches can be viewed as stylized edge maps, or many artistic styles merely impose brushstroke patterns and color palettes onto the input image. Stylized abstraction, however, demands more than stylistic transfer; it requires a careful balance between simplification and recognizability. It involves distilling the subject to its most \textit{iconic traits}, sometimes exaggerating them while discarding fine-grained details. This abstraction introduces greater semantic and structural deviation from the input, making the problem far more complex than merely applying texture or color-based transformations.

Now, while a number of image stylization methods exist ranging from training-free techniques to fine-tuning-based pipelines \cite{ruiz2023dreambooth, gal2022image} or encoder-based methods \cite{li2023blip, ye2023ip} that adapt reference features for concept preservation in general T2I models these approaches often fall short in the domain of stylized abstraction. Notably, many existing works demonstrate results primarily on celebrity faces, which are already well-represented in pre-trained models. As a result, these models often succeed in retaining recognizable features simply because they have been exposed to those identities during training. However, when tested on images of everyday individuals, these same methods either fail to preserve identity or compromise the intended stylization (Visualized in Figure \ref{fig:teaser} (b)). This highlights the need for methods that can generalize stylized abstraction to diverse identities without relying on prior memorization.


To address the limitations of existing stylization methods, we present a training-free framework for stylized abstraction that generalizes beyond celebrity likenesses to everyday identities and supports a wide range of abstract styles: including LEGO, South Park, Simpson, Matrushka, Barbie, Knitted Doll, Action Figure, etc. Our approach requires neither subject-specific fine-tuning nor dataset-level adaptation. Instead, we introduce a novel inference-time scaling strategy for vision-language models (VLLMs) that distills core semantic traits critical for identity preservation and aligns them with user-driven stylistic prompts. Central to our pipeline is a multi-turn generation loop, where missing or distorted identity cues, identified via VLLM feedback, are progressively reintegrated to enhance fidelity across iterations. To recover subject structure under extreme abstraction, we extend RF-Inversion \cite{rout2025semantic} with a cross-domain latent inversion scheme, treating stylized images as source latents and photorealistic representations as structural targets. Leveraging rectified flow-guided updates and style-aware temporal scheduling, our method preserves stylistic fidelity while selectively restoring identity-consistent structure in a controllable and interpretable fashion. To evaluate abstraction beyond pixel-level similarity, we introduce StyleBench, a GPT-assisted, human-aligned protocol for benchmarking stylized abstraction. We further report quantitative performance using KID and CLIPScore, supported by a user preference study. Our framework sets a new state-of-the-art in abstraction quality: fully training-free, identity-consistent, and broadly generalizable across styles and subjects.


\section{Related Work}
\noindent \textbf{Identity-Preserving Style Transfer.}  Identity-preserving or subject-driven style transfer \cite{zhang2024ssr,raj2023dreambooth3d, chen2023subject,miao2024subject,dong2022dreamartist, han2023highly,voynov2023p+,alaluf2023neural,kumari2023multi,liu2023cones,han2023svdiff,ryu2023low,avrahami2023break, chen2023disenbooth,cai2024decoupled,he2025data} is a closely related line of work to stylized abstraction, where the goal is to synthesize stylized images while retaining subject identity. Approaches in this domain broadly fall into three categories. The first category includes fine-tuning-based methods, such as Textual Inversion~\cite{gal2022image} and DreamBooth~\cite{ruiz2023dreambooth}, which adapt the generative model to the target subject using multiple reference images. While effective for object-centric domains, these methods often struggle to faithfully preserve identity in human subjects and require several input images for fine-tuning. The second category comprises encoder-based methods that learn feature adaptation modules to modulate the base model without explicit fine-tuning. Notable examples include IP-Adapter~\cite{ye2023ip} and BLIP-Diffusion~\cite{li2023blip}, which leverage pretrained encoders to align content and style representations. The third category focuses on single-image personalization, including DreamTuner~\cite{hua2023dreamtuner} and CSGO~\cite{xing2024csgo}, or entirely training-free techniques such as RF-Inversion~\cite{rout2025semantic}, RB-Modulation~\cite{rout2025rbmodulation}, StyleID~\cite{le2022styleid}, InstantID~\cite{wang2024instantid}, InstantID-Plus~\cite{wang2024instantstyle}, and DiffArtist~\cite{jiang2024diffartist}. These methods often rely on CLIP-guided optimization or feature injection to steer generation toward the desired style. However, such methods typically prioritize structural fidelity or stylization strength in isolation. In high stylization scenarios, they may struggle to abstract and reinterpret content meaningfully, as their architectures tend to enforce either identity preservation or stylistic consistency without deeper semantic understanding. Bridging this gap requires novel approaches that reason over semantic correspondences between style and identity, rather than relying solely on pixel or feature-space alignment.

\noindent \textbf{Multi-Modal LLMs in Personalized Image Generation.}  
Recent advances in multi-modal large language models (MLLMs) have demonstrated their potential in various image generation tasks~\cite{gal2022image,liu2025llm4gen,liao2024text,wu2024self,sun2024autoregressive,wu2024next}, although not always directly targeting personalized image generation. These models exhibit strong generalization capabilities when applied to complex and previously unseen scenarios~\cite{hu2024instruct, qu2023layoutllm, wang2024genartist}. Leveraging both multi-modal understanding and generative modeling, commercial models such as GPT-4o \cite{openai2024gpt4o}, Gemini \cite{deepmind2023gemini}, and Grok \cite{xai2025grok} have recently shown the ability to produce stylized and personalized images from user inputs. However, we highlight several limitations in this emerging line of work. Most of these systems are proprietary and closed-source, trained on large-scale datasets that are not publicly available. Furthermore, it remains unclear whether the outputs involve additional fine-tuning or personalization modules beyond the core model. These factors hinder reproducibility and limit academic scrutiny.



\section{Method}

\begin{figure}[ht]
    \centering
\includegraphics[width=1\textwidth]{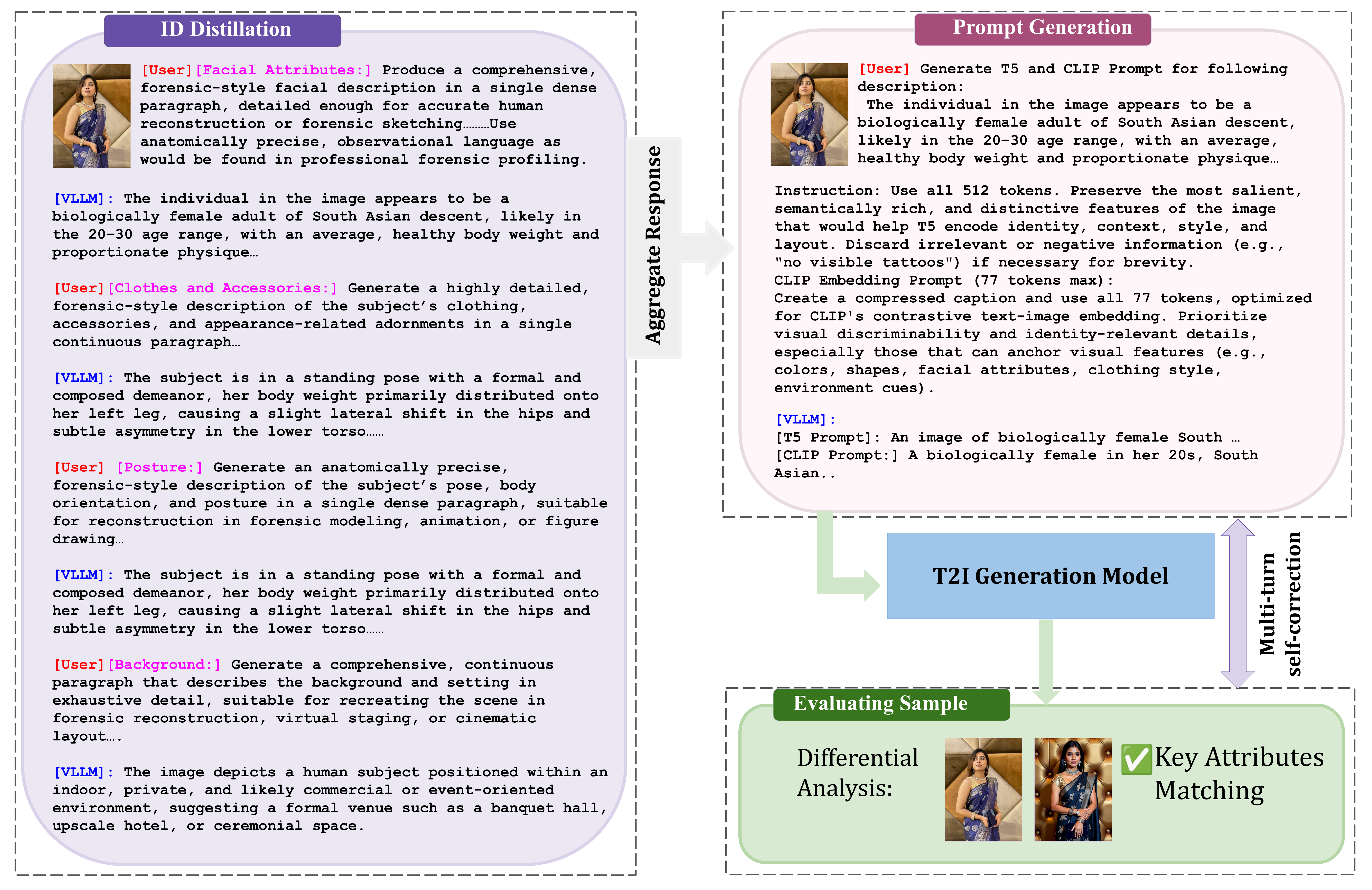}
    \caption{\textbf{Workflow of identity distillation via inference-time VLLM scaling.} The process includes dense attribute extraction, multi-scale prompt compression, iterative identity refinement, and style-aware prompt transformation.}
    \label{fig:method}
\end{figure}

\subsection{Subject Identity Distillation via Inference-Time VLLM Scaling}

\noindent \textbf{Dense Attribute Extraction.}
Given an input image $\mathbf{I} \in \mathbb{R}^{H \times W \times 3}$, we initiate a multi-round interaction with a vision-language language model (VLLM) \cite{zhu2025internvl3} to obtain exhaustive descriptions of identity-related features. Let $\mathcal{V}$ denote the VLLM interface. The process is divided into four semantically disjoint rounds: facial attributes ($\mathcal{A}_\text{face}$), clothing and accessories ($\mathcal{A}_\text{attire}$), posture and pose ($\mathcal{A}_\text{pose}$), and background environment ($\mathcal{A}_\text{scene}$). Each round is conditioned on $\mathbf{I}$ and prompts $\mathcal{P}_k$ specifically tailored to query salient features of category $k$:

\begin{equation}
    \mathcal{A}_k = \mathcal{V}(\mathbf{I}, \mathcal{P}_k), \quad \text{for } k \in \{\text{face}, \text{attire}, \text{pose}, \text{scene}\}.
\end{equation}
\noindent The outputs $\{\mathcal{A}_k\}$ are structured natural language descriptions optimized for visual grounding.

\noindent \textbf{Multi-Scale Prompt Compression.}
The extracted descriptions are aggregated and passed to a secondary VLLM instance $\mathcal{V}'$, which synthesizes two task-specific prompts- a) $\mathcal{T}_{512}$: A 512-token prompt optimized for T5-based \cite{raffel2020exploring} generators, preserving identity, context, style, and layout. b) $\mathcal{T}_{77}$: A 77-token CLIP-style \cite{radford2021learning} prompt, distilled to maximize contrastive relevance in embedding space.

\noindent Formally, let $\mathcal{A}_{\text{full}} = \bigcup_k \mathcal{A}_k$, then:
\begin{equation}
    \mathcal{T}_{512}, \mathcal{T}_{77} = \mathcal{V}'(\mathcal{A}_{\text{full}}).
\end{equation}

\noindent \textbf{Iterative Identity Refinement.}
The condensed prompts $\mathcal{T}_{512}, \mathcal{T}_{77}$ are used as conditioning inputs to an image generation pipeline, Flux \cite{flux2024}, producing a candidate image $\hat{\mathbf{I}}$. A third VLLM instance $\mathcal{V}''$ performs a differential analysis between the original image $\mathbf{I}$ and generated image $\hat{\mathbf{I}}$, identifying missing or misaligned attributes:

\begin{equation}
    \Delta \mathcal{A} = \mathcal{V}''(\mathbf{I}, \hat{\mathbf{I}}).
\end{equation}

\noindent These attributes $\Delta \mathcal{A}$ are incrementally reintegrated into the textual representation by updating $\mathcal{A}_{\text{full}} \leftarrow \mathcal{A}_{\text{full}} \cup \Delta \mathcal{A}$, prompting a regeneration of $\mathcal{T}_{512}, \mathcal{T}_{77}$. This loop continues until either a perceptual alignment threshold is reached (e.g., $\text{CLIP}(\mathbf{I}, \hat{\mathbf{I}}) \geq \tau$) or a maximum number of rounds $T$ is completed.

\noindent \textbf{Inference-Time Identity Convergence.}
This inference-time distillation strategy enables progressive identity preservation without requiring gradient updates. The architecture remains fixed; only VLLM feedback adaptively steers prompt construction. The convergence criterion is defined as:
\begin{equation}
    \text{stop} \iff \text{CLIP}(\mathbf{I}, \hat{\mathbf{I}}^{(t)}) \geq \tau \quad \text{or} \quad t \geq T,
\end{equation}
where $\hat{\mathbf{I}}^{(t)}$ is the generated image at iteration $t$.

\noindent This multi-turn VLLM-in-the-loop mechanism emulates a self-correcting distillation process, bridging perceptual gaps between the source image and its identity representation without paired supervision.

\noindent \textbf{Style-Aware Prompt Transformation.}
The updated prompt pair $(\mathcal{T}{512}, \mathcal{T}{77})$ undergoes a style-conditioning step. A final VLLM module $\mathcal{V}^\ast$ is invoked with a style descriptor $\mathcal{S}$ (e.g., "knitted doll", "LEGO", or "anime") to adapt the identity-rich prompt into a stylized version while preserving semantic fidelity:

\begin{equation}
\mathcal{T}^{\text{styled}}{512}, \mathcal{T}^{\text{styled}}{77} = \mathcal{V}^\ast(\mathcal{T}{512}, \mathcal{T}{77}, \mathcal{S}).
\end{equation}

\noindent These stylized prompts guide the Flux generation pipeline to produce an initial abstraction $y_s$. An overview of the framework is shown in Figure~\ref{fig:method}.

\subsection{Cross-Domain Latent Reversal with Rectified Flows}

At this stage, we obtain a highly stylized representation of the subject with faithfully preserved stylistic elements, $y_s$, but the generation now requires structural guidance from the original image to recover key identity-aligned geometry. However, unlike prior inversion-based pipelines \cite{rout2025semantic} that operate on realistic or noisy inputs, our setting begins from an already abstracted, stylized image $y_s$. This shift introduces a novel challenge: how to reconstruct a semantically grounded latent representation from a highly altered input while flexibly recovering structural details based on style demands.

To address this, we propose a two-stage framework for cross-domain latent reversal, which extends rectified flow methods \cite{rout2025semantic} into the abstract stylization space. The key is to treat the stylized abstraction not as a degraded variant of a photo, but as a valid starting point in an altered visual domain: one that must be softly regularized back into a structured latent aligned with the true subject identity.

In the first stage, we invert the stylized image $y_s$ using a forward rectified ODE:

\begin{equation}
dY_t = \left[ u_t(Y_t) + \gamma \left( u_t(Y_t \mid y_1) - u_t(Y_t) \right) \right] dt, \quad Y_0 = y_s,
\end{equation}

where $u_t(Y_t)$ denotes the unconditional drift field from the pretrained Flux model, and $u_t(Y_t \mid y_1)$ is an analytically derived controller via linear quadratic regulation (LQR). The scalar $\gamma \in [0, 1]$ governs the balance between staying close to the stylized abstraction and conforming to the learned noise prior $y_1 \sim \mathcal{N}(0, I)$. This inversion step introduces the application of rectified flows to high-level stylistic abstractions, where the latent does not correspond to a photo-realistic image but to a semantically enriched domain-specific representation.

The second stage performs structure-aware reconstruction using a controlled reverse ODE, initialized from $y_1$ and guided toward a real reference image $y_r$:

\begin{equation}
dX_t = \left[ v_t(X_t) + \eta_t \left( v_t(X_t \mid y_r) - v_t(X_t) \right) \right] dt, \quad X_0 = y_1,
\end{equation}

where $v_t$ is the reverse-time vector field. The time-varying structural controller $\eta_t$ defined as:

\begin{equation}
\eta_t = 
\begin{cases}
\eta, & t \in [\tau_{\text{start}}, \tau_{\text{stop}}] \\
0, & \text{otherwise}
\end{cases}
\end{equation}

\begin{figure*}[!htb]
    \centering
    \includegraphics[width=1\textwidth]{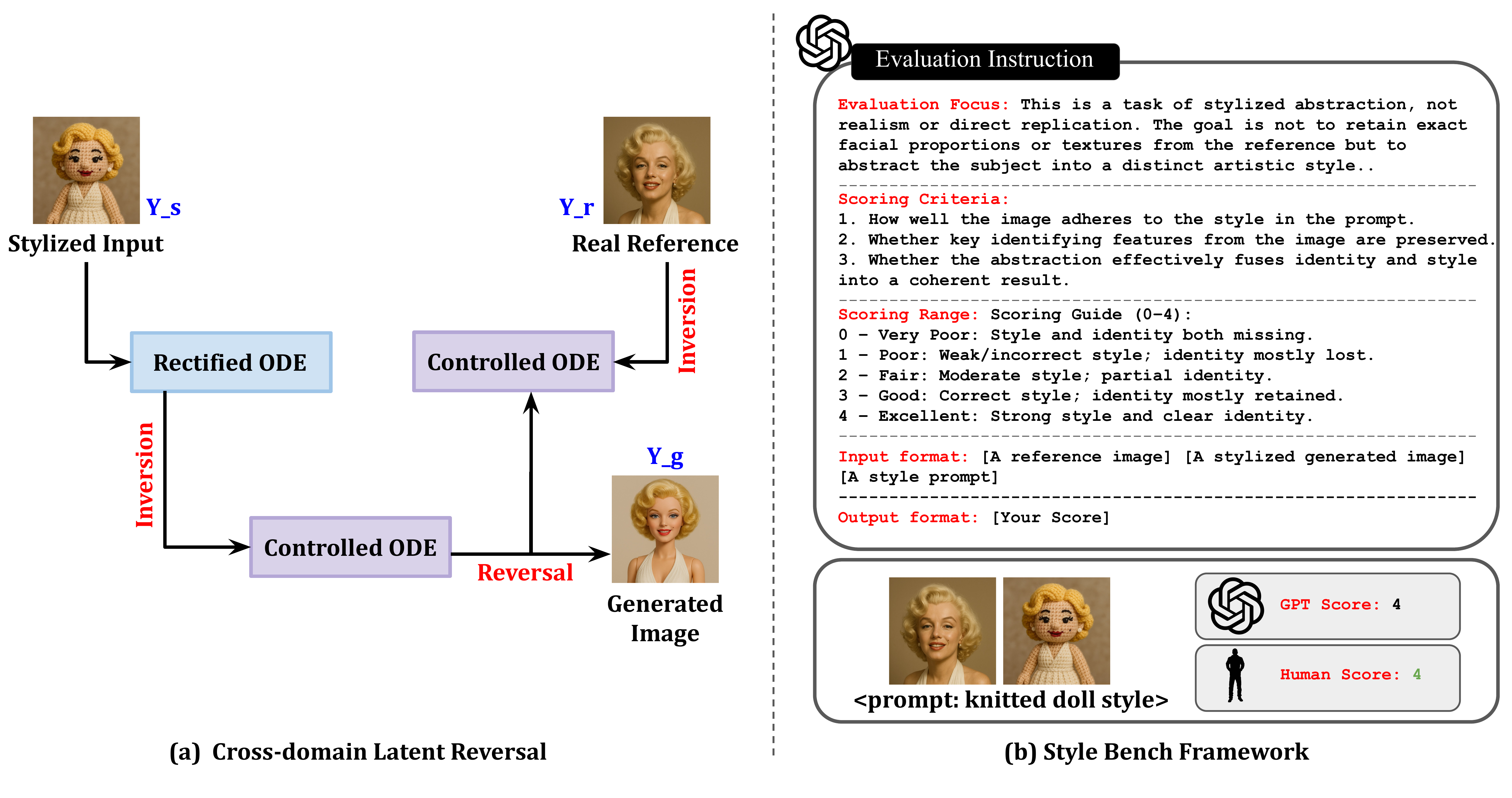}
    \vspace{-20pt}
    \caption{ (a) Cross-domain Latent Reversal pipeline for stylized image generation.
(b) End-to-end workflow of the StyleBench evaluation framework.}
    \label{fig:cross-dom-method}
    \vspace{-10pt}
\end{figure*}

Unlike fixed-strength guidance, this scheduling allows us to inject structural constraints only during a \textbf{style-dependent} temporal window. Such a design is critical for abstract styles where early over-regularization can collapse style integrity. Parameters $\eta$, $\tau_{\text{start}}$, and $\tau_{\text{stop}}$ are adaptively chosen using a VLLM-based controller that parses the style descriptor (e.g. "knitted doll", "South Park") to determine the structural necessity. An overview of the process is shown in Figure~\ref{fig:cross-dom-method} (a).

\subsection{StyleBench: Human-Aligned Evaluation for Stylized Abstraction}

Evaluating stylized image generation requires going beyond traditional notions of visual similarity. In many artistic or character-driven styles, such as knitted dolls, LEGO figures, or South Park characters, the geometry, texture, and proportions of the original subject are intentionally distorted. However, what remains essential is the preservation of key identity cues: hairstyle, posture, clothing, or accessories that allow recognition despite abstraction. Existing benchmarks like DreamBench++\cite{peng2024dreambench++} have made progress in human-aligned evaluation for personalized image generation, particularly by assessing prompt consistency and subject fidelity using multimodal GPT-4o model. However, these benchmarks primarily operate under assumptions of semantic and structural coherence typical of photorealistic or lightly stylized domains.

In contrast, \textit{StyleBench} is tailored specifically for \textit{stylized abstraction}, where the visual transformation is often extreme and the identity must be reinterpreted through a unique stylistic lens. Our benchmark introduces a structured evaluation protocol using GPT models, which are prompted with three inputs: a reference image, a stylized generation, and a style prompt. The task definition guides the model to assess not realism or one-to-one replication, but how well the abstraction balances fidelity to the subject's recognizable identity with faithful adherence to the style’s visual language. 

To ensure consistent and human-aligned evaluation, we design the GPT prompt with explicit scoring criteria across three integrated axes: (i) adherence to style, (ii) identity preservation, and (iii) fusion quality. The evaluation process incorporates internal task summarization and optional chain-of-thought reasoning to encourage self-alignment \cite{peng2024dreambench++,sun2023principle} before issuing a score between 0 (very poor) and 4 (excellent). Unlike generic perceptual metrics (e.g., CLIP \cite{radford2021learning}, DINO \cite{zhang2022dino}), which often fail under stylization shifts, our protocol enables nuanced judgments aligned with how humans interpret abstracted identity. This makes \textit{StyleBench} particularly suitable for benchmarking models that target stylized avatars, artistic reinterpretations, and toy-based renderings, domains where abstraction is not a flaw but a defining feature.

\vspace{-10pt}
\section{Experiment}


\textbf{Baselines}
There is no direct baseline for stylized abstraction, as it is a relatively new concept in the vision community that goes beyond identity preservation to include semantic and geometric reinterpretation. We compare against the closest related methods across personalization and style transfer. These include fine-tuning-based approaches such as Textual Inversion~\cite{gal2022image} and DreamBooth~\cite{ruiz2023dreambooth}, encoder-based CSGO~\cite{xing2024csgo}, and training-free, zero-shot methods including StyleID~\cite{le2022styleid}, RF-Inversion~\cite{rout2025semantic}, RB-Modulation~\cite{rout2025rbmodulation}, DiffArtist~\cite{jiang2024diffartist}, InstantID~\cite{wang2024instantid}, and InstantID-Plus~\cite{wang2024instantstyle}.


\textbf{Dataset and Evaluation metric.} Our dataset consists of three categories: (i) single subject images of everyday individuals, comprising 10 images across 10 unique subjects; (ii) multi-subject images of everyday individuals, totaling 14 images; and (iii) single-subject celebrity images collected from Google Image Search under free use licenses, amounting to 30 images.

For evaluation, we employ Kernel Inception Distance (KID) \cite{binkowski2018demystifying}, CLIP score, our proposed StyleBench benchmark, and human evaluation. The human evaluation is conducted on 25 generated images, rated by 15 independent annotators.

\textbf{Implementation Details}
We implement all models using PyTorch and run experiments on NVIDIA A6000. We employ InternVL \cite{zhu2025internvl3} as the VLLM and FLUX \cite{flux2024} as the image generation backbone. More details on baseline preprocessing and reproduction are provided in the Appendix.

\section{Results and Analysis}
\begin{figure*}[!htb]
    \centering
    \includegraphics[width=1\textwidth]{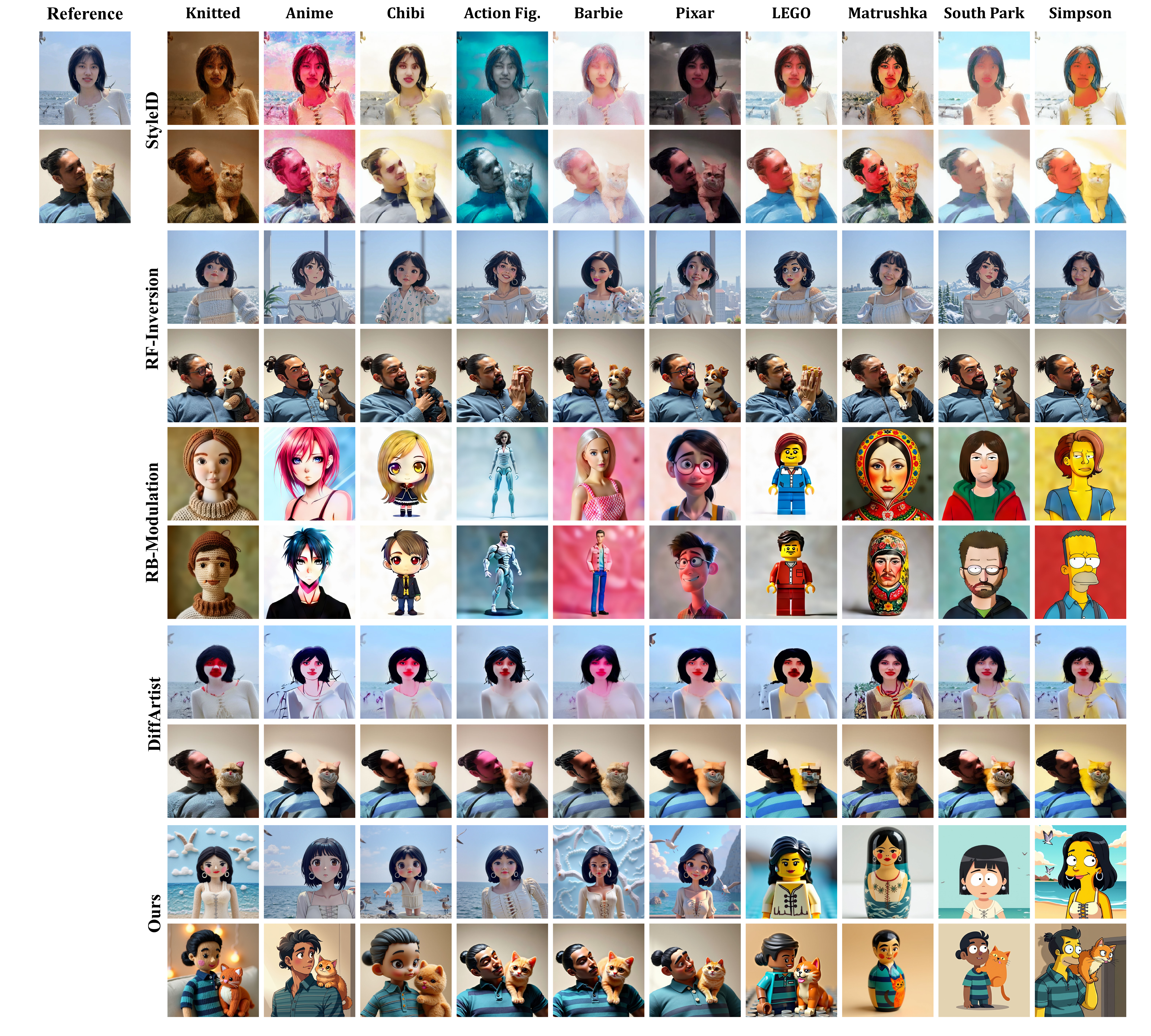}
    \vspace{ -15pt}
    \caption{\textbf{Qualitative comparison with existing image stylization models.} Most prior methods struggle to preserve either the reference content or the intended style. For example, models like StyleID \cite{le2022styleid} rely on a reference style image and often perform low-level pixel blending, which fails to generalize to high-level abstractions. In contrast, our method preserves both identity and style with high semantic fidelity.
}
    \label{fig:main-qual}
\end{figure*}

\begin{figure*}[!htb]
    \centering
    \includegraphics[width=1\textwidth]{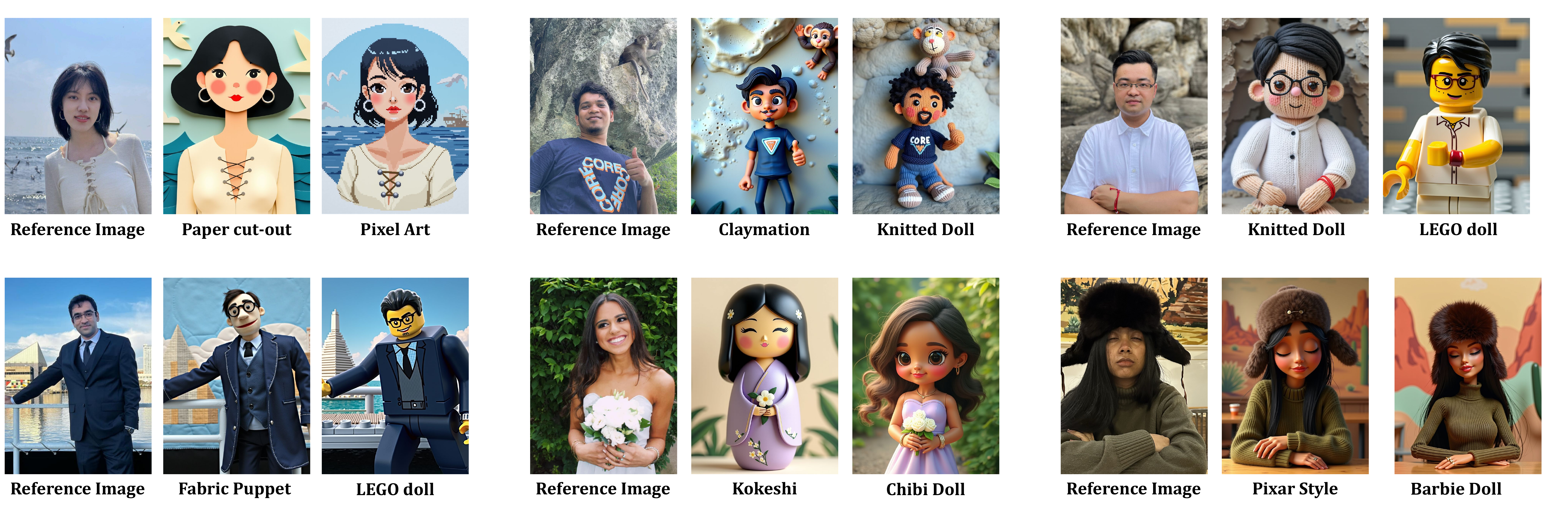}
    \vspace{-10pt}
    \caption{\textbf{Additional results across diverse subjects and abstract styles.}}
    \vspace{-5pt}
    \label{fig:additionals}
\end{figure*}

\textbf{Qualitative Results.} Figure~\ref{fig:main-qual} presents a diverse set of stylized abstractions across 10 styles, including regional representations and balanced gender coverage. Baseline methods often fail because they are not specifically designed for high-level abstraction tasks, particularly in training-free settings with only a single reference image. These models typically lack mechanisms to semantically disentangle identity from style, leading to poor content preservation or shallow stylistic transfer. On the other hand, our method consistently retains subject essence while embracing the stylistic exaggeration unique to each domain. Some additional results with more concepts are shown in Figure \ref{fig:additionals}.

\textbf{Quantitative Results.} Table~\ref{tab:style_comparison_grouped} compares our method against existing baselines across KID, CLIP score, StyleBench, and human evaluation. Our method achieves competitive performance across all metrics, particularly excelling in human-aligned scores, highlighting its ability to produce abstractions that are both recognizable and stylistically faithful.

\begin{table*}[t]
    \centering
    \caption{Comparison of stylization methods across KID, CLIP, StyleBench, and human evaluation scores. Methods are grouped into fine-tuned, encoder-based, and training-free categories.}
    \resizebox{1\textwidth}{!}{%
    \begin{tabular}{llcccc}
        \toprule
        \textbf{Category} & \textbf{Method} & \textbf{KID $\downarrow$} & \textbf{CLIP Score $\uparrow$} & \textbf{StyleBench $\uparrow$} & \textbf{Human Eval $\uparrow$} \\
        \midrule
        \multirow{2}{*}{\textbf{Fine-tuned}} 
            & Textual Inversion~\cite{gal2022image} & 0.042 & 0.2124 &  0 & 0.5 \\
            & DreamBooth~\cite{ruiz2023dreambooth}     & 0.036 & 0.1910 & 0 & 1 \\
        \midrule
        \textbf{Encoder-based} 
            & CSGO~\cite{xing2024csgo}           & 0.140 & 0.1977 & 1.5 & 1 \\
        \midrule
        \multirow{7}{*}{\textbf{Training-Free}} 
            & StyleID~\cite{le2022styleid}        & 0.213 & 0.2161 & 1.5 & 1.5 \\
            & RF-Inversion~\cite{rout2025semantic}   & 0.166 & 0.1902 & 1.5 & 2 \\
            & RB-Modulation~\cite{rout2025rbmodulation}  & 0.035 & 0.2069 & 0.5 & 0.5 \\
            & DiffArtist~\cite{jiang2024diffartist}     & 0.255 & 0.1966 & 1.75 & 0.5 \\
            & InstantID~\cite{wang2024instantid}      & 0.035 & 0.2168 & 1 & 1.5 \\
            &  \textbf{Ours}           & \textbf{0.025} & \textbf{0.2272} & \textbf{4} & \textbf{3.8} \\
        \bottomrule
    \end{tabular}}
    \vspace{-10pt}
    \label{tab:style_comparison_grouped}
\end{table*}

\begin{table*}[!ht]
    \centering
    \caption{CLIP scores under different feedback conditions. Evaluation is done with the same reference across all prompt stages.}
    \begin{tabular}{lccccc}
        \toprule
        \textbf{Prompt Type} & \textbf{Vanilla Prompt} & \textbf{Feedback-1} & \textbf{Feedback-2} & \textbf{Feedback-3} & \textbf{Verifier} \\
        \midrule
        CLIP Score & 0.6558 & 0.6980 &  0.7367  & 0.8494 & 0.8575 \\
        \bottomrule
    \end{tabular}
    \label{tab:clip_feedback}
    \vspace{-5pt}
\end{table*}

\begin{figure*}[!htb]
    \centering
    \includegraphics[width=1\textwidth]{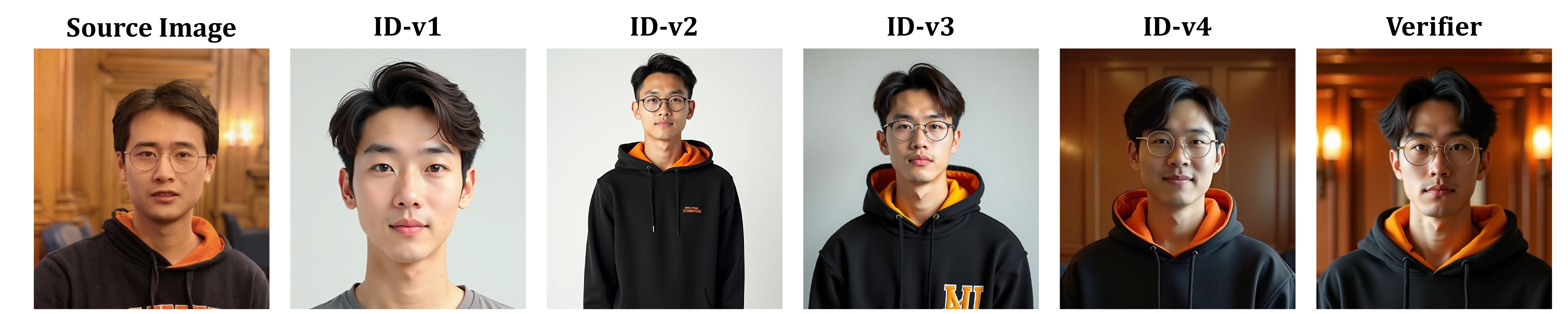}
    \vspace{-15pt}
    \caption{\textbf{Multi-round inference-time scaling with VLLMs for identity distillation.} At each round, the VLLM extracts identity-relevant features from the original image to reconstruct a refined base representation. This iterative process progressively distills semantic identity (e.g., facial structure, clothing, posture) while filtering out irrelevant details. The final distilled output serves as a robust foundation for stylized abstraction, enabling faithful and expressive generation across diverse styles.}
    \label{fig:inf-scale}
\end{figure*}

\textbf{Impact of Identity-Distilled Prompts.} We investigate the effect of using dense, identity-distilled prompts obtained via inference-time querying of a VLLM to extract subject-specific attributes. This experiment evaluates how effectively the model can reconstruct the original image when conditioned solely on the prompt derived from that image. Table~\ref{tab:clip_feedback} reports CLIP scores under different feedback conditions, highlighting the influence of inference-time scaling on identity fidelity. Qualitative examples are shown in Figure~\ref{fig:inf-scale}.

\begin{figure*}[!htb]
    \centering
    \includegraphics[width=1\textwidth]{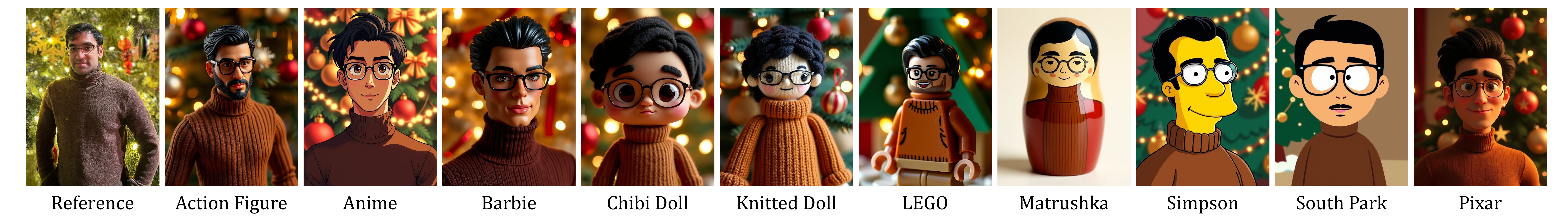}
    \vspace{-10pt}
    \caption{ \textbf{Stylized Generation from text-only Prompts after Identity Distillation.} In this stage, the image is no longer used, only the distilled stylized text prompt is fed to the image generation model. The resulting stylized outputs preserve key identity traits such as hairstyle, clothing, and pose, despite the absence of direct visual reference. This demonstrates the effectiveness of our identity distillation pipeline in guiding style-consistent abstraction purely from text.}
    \vspace{-5pt}
    \label{fig:prompt-style}
    
\end{figure*}

\textbf{Impact of Prompt Stylization.} Prompt stylization involves enriching the original prompt with style-consistent descriptors, for e.g., replacing generic phrases with detailed attributes such as "button eyes" or "yarn hair" for knitted dolls, or "yellow skin" for Simpsons-style characters. This guides the model toward more faithful stylistic abstraction.  Qualitative differences are shown in Figure~\ref{fig:prompt-style}.

\begin{figure*}[!htb]
    \centering
    \vspace{-5pt}
    \includegraphics[width=1\textwidth]{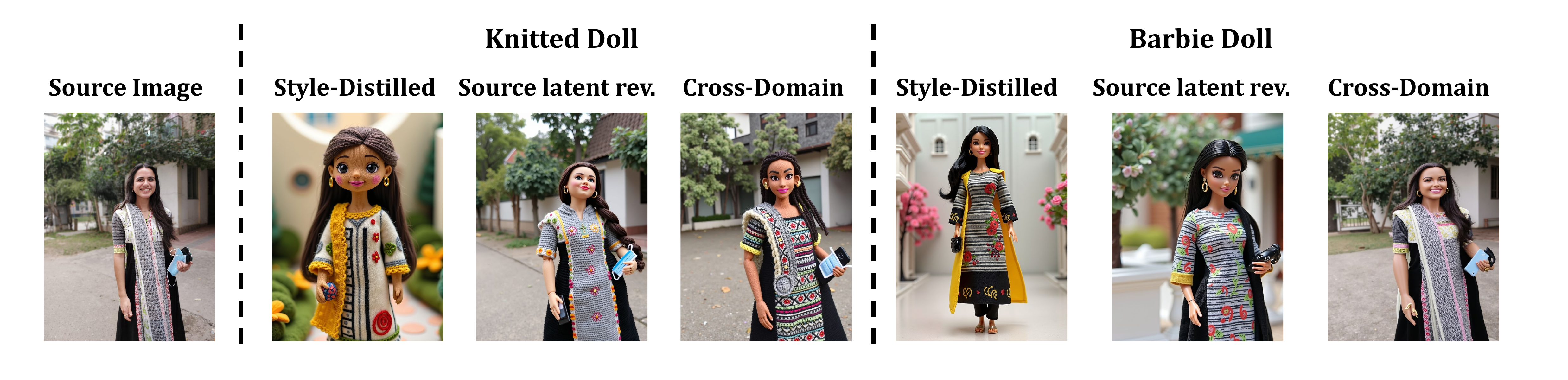}
    \vspace{-15pt}
    \caption{\textbf{Effect of Cross-Domain Latent Reversal.} Given a stylized reference and original image, our method uses a VLLM to balance style and structure. Cross-domain reversal starts from a text-initialized latent and iteratively aligns with the reference style while preserving structure. In contrast, source latent reversal starts from the original and applies the style prompt, often disrupting structure. Our approach yields more coherent, identity-preserving abstractions.
.}
    \vspace{-5pt}
    \label{fig:cross-domain}
    \vspace{-5pt}
\end{figure*}

\textbf{Impact of Cross-Domain Latent Reversal.}
To evaluate the effectiveness of our cross-domain latent reversal framework, we present qualitative comparisons in Figure \ref{fig:cross-domain}. The source latent reversal baseline starts from the original image and uses a densely styled prompt to directly generate a stylized output. However, this approach often fails to capture details such as in the knitted doll example, the facial texture lacks the distinctive knitted pattern due to the absence of a strongly stylized starting point. In contrast, our cross-domain latent reversal begins from an already stylized abstraction, resulting in better preservation of style-specific features. Similarly, in the Barbie doll case, source latent reversal over-emphasizes style at the cost of structural integrity, while our method achieves a more balanced reconstruction of both style and identity.

\section{Conclusion}
We present a training-free framework for stylized abstraction that integrates vision-language inference scaling with cross-domain rectified flow inversion strategy. By dynamically modulating structural restoration based on learned style priors, our method enables faithful yet flexible abstraction across a wide range of stylized domains. Combined with a new evaluation protocol, \textit{StyleBench}, this work establishes a foundation for abstraction-aware generation of everyday subjects, supporting creative applications without requiring model fine-tuning.

\textbf{Limitations} Our method inherits limitations from the underlying VLLM and image generation models, particularly in handling rare styles and edge cases, which may impact output fidelity and generalization. Notably, the model can exhibit racial or cultural biases, such as consistently associating "South Asian" prompts with bindis or traditional jewelry, or "Middle Eastern" with facial hair, regardless of whether these features are present in the reference image. These biases reflect broader challenges in mitigating race-related stereotyping within generative models.

 \textbf{Ethical Statement} We use real human images with explicit consent strictly for research purposes. This work supports applications in areas like merchandising, creative content generation, and ideation. Any potential misuse is against our intentions and values.


\bibliography{egbib}

\begin{thebibliography}{10}

\bibitem{alaluf2023neural}
Yuval Alaluf, Elad Richardson, Gal Metzer, and Daniel Cohen-Or.
\newblock A neural space-time representation for text-to-image personalization.
\newblock {\em ACM Transactions on Graphics (TOG)}, 42(6):1--10, 2023.

\bibitem{avrahami2023break}
Omri Avrahami, Kfir Aberman, Ohad Fried, Daniel Cohen-Or, and Dani Lischinski.
\newblock Break-a-scene: Extracting multiple concepts from a single image.
\newblock In {\em SIGGRAPH Asia 2023 Conference Papers}, pages 1--12, 2023.

\bibitem{berger2013style}
Itamar Berger, Ariel Shamir, Moshe Mahler, Elizabeth Carter, and Jessica Hodgins.
\newblock Style and abstraction in portrait sketching.
\newblock {\em ACM Transactions on Graphics (TOG)}, 32(4):1--12, 2013.

\bibitem{binkowski2018demystifying}
Miko{\l}aj Bi{\'n}kowski, Danica~J Sutherland, Michael Arbel, and Arthur Gretton.
\newblock Demystifying mmd gans.
\newblock {\em arXiv preprint arXiv:1801.01401}, 2018.

\bibitem{cai2024decoupled}
Yufei Cai, Yuxiang Wei, Zhilong Ji, Jinfeng Bai, Hu~Han, and Wangmeng Zuo.
\newblock Decoupled textual embeddings for customized image generation.
\newblock In {\em Proceedings of the AAAI Conference on Artificial Intelligence}, volume~38, pages 909--917, 2024.

\bibitem{chen2024artadapter}
Dar-Yen Chen, Hamish Tennent, and Ching-Wen Hsu.
\newblock Artadapter: Text-to-image style transfer using multi-level style encoder and explicit adaptation.
\newblock In {\em Proceedings of the IEEE/CVF conference on computer vision and pattern recognition}, pages 8619--8628, 2024.

\bibitem{chen2021artistic}
Haibo Chen, Zhizhong Wang, Huiming Zhang, Zhiwen Zuo, Ailin Li, Wei Xing, Dongming Lu, et~al.
\newblock Artistic style transfer with internal-external learning and contrastive learning.
\newblock {\em Advances in Neural Information Processing Systems}, 34:26561--26573, 2021.

\bibitem{chen2023disenbooth}
Hong Chen, Yipeng Zhang, Xin Wang, Xuguang Duan, Yuwei Zhou, and Wenwu Zhu.
\newblock Disenbooth: Disentangled parameter-efficient tuning for subject-driven text-to-image generation.
\newblock {\em arXiv preprint arXiv:2305.03374}, 3(4), 2023.

\bibitem{chen2023subject}
Wenhu Chen, Hexiang Hu, Yandong Li, Nataniel Ruiz, Xuhui Jia, Ming-Wei Chang, and William~W Cohen.
\newblock Subject-driven text-to-image generation via apprenticeship learning.
\newblock {\em Advances in Neural Information Processing Systems}, 36:30286--30305, 2023.

\bibitem{deepmind2023gemini}
Google DeepMind.
\newblock Gemini: A family of highly capable multimodal models.
\newblock \url{https://arxiv.org/abs/2312.11805}, 2023.
\newblock Accessed: 2025-05-10.

\bibitem{deng2022stytr2}
Yingying Deng, Fan Tang, Weiming Dong, Chongyang Ma, Xingjia Pan, Lei Wang, and Changsheng Xu.
\newblock Stytr2: Image style transfer with transformers.
\newblock In {\em Proceedings of the IEEE/CVF conference on computer vision and pattern recognition}, pages 11326--11336, 2022.

\bibitem{dong2022dreamartist}
Ziyi Dong, Pengxu Wei, and Liang Lin.
\newblock Dreamartist: Towards controllable one-shot text-to-image generation via positive-negative prompt-tuning.
\newblock {\em arXiv preprint arXiv:2211.11337}, 2022.

\bibitem{gal2022image}
Rinon Gal, Yuval Alaluf, Yuval Atzmon, Or~Patashnik, Amit~H Bermano, Gal Chechik, and Daniel Cohen-Or.
\newblock An image is worth one word: Personalizing text-to-image generation using textual inversion.
\newblock {\em arXiv preprint arXiv:2208.01618}, 2022.

\bibitem{gatys2015neural}
Leon~A Gatys, Alexander~S Ecker, and Matthias Bethge.
\newblock A neural algorithm of artistic style.
\newblock {\em arXiv preprint arXiv:1508.06576}, 2015.

\bibitem{han2023highly}
Inhwa Han, Serin Yang, Taesung Kwon, and Jong~Chul Ye.
\newblock Highly personalized text embedding for image manipulation by stable diffusion.
\newblock {\em arXiv preprint arXiv:2303.08767}, 2023.

\bibitem{han2023svdiff}
Ligong Han, Yinxiao Li, Han Zhang, Peyman Milanfar, Dimitris Metaxas, and Feng Yang.
\newblock Svdiff: Compact parameter space for diffusion fine-tuning.
\newblock In {\em Proceedings of the IEEE/CVF International Conference on Computer Vision}, pages 7323--7334, 2023.

\bibitem{he2025data}
Xingzhe He, Zhiwen Cao, Nicholas Kolkin, Lantao Yu, Kun Wan, Helge Rhodin, and Ratheesh Kalarot.
\newblock A data perspective on enhanced identity preservation for diffusion personalization.
\newblock In {\em 2025 IEEE/CVF Winter Conference on Applications of Computer Vision (WACV)}, pages 3782--3791. IEEE, 2025.

\bibitem{hu2024instruct}
Hexiang Hu, Kelvin~CK Chan, Yu-Chuan Su, Wenhu Chen, Yandong Li, Kihyuk Sohn, Yang Zhao, Xue Ben, Boqing Gong, William Cohen, et~al.
\newblock Instruct-imagen: Image generation with multi-modal instruction.
\newblock In {\em Proceedings of the IEEE/CVF conference on computer vision and pattern recognition}, pages 4754--4763, 2024.

\bibitem{hua2023dreamtuner}
Miao Hua, Jiawei Liu, Fei Ding, Wei Liu, Jie Wu, and Qian He.
\newblock Dreamtuner: Single image is enough for subject-driven generation.
\newblock {\em arXiv preprint arXiv:2312.13691}, 2023.

\bibitem{jiang2024diffartist}
Ruixiang Jiang and Changwen Chen.
\newblock Diffartist: Towards structure and appearance controllable image stylization, 2024.

\bibitem{jing2019neural}
Yongcheng Jing, Yezhou Yang, Zunlei Feng, Jingwen Ye, Yizhou Yu, and Mingli Song.
\newblock Neural style transfer: A review.
\newblock {\em IEEE transactions on visualization and computer graphics}, 26(11):3365--3385, 2019.

\bibitem{kumari2023multi}
Nupur Kumari, Bingliang Zhang, Richard Zhang, Eli Shechtman, and Jun-Yan Zhu.
\newblock Multi-concept customization of text-to-image diffusion.
\newblock In {\em Proceedings of the IEEE/CVF conference on computer vision and pattern recognition}, pages 1931--1941, 2023.

\bibitem{flux2024}
Black~Forest Labs.
\newblock Flux.
\newblock \url{https://github.com/black-forest-labs/flux}, 2024.

\bibitem{le2022styleid}
Minh-Ha Le and Niklas Carlsson.
\newblock Styleid: Identity disentanglement for anonymizing faces.
\newblock {\em arXiv preprint arXiv:2212.13791}, 2022.

\bibitem{li2023blip}
Dongxu Li, Junnan Li, and Steven Hoi.
\newblock Blip-diffusion: Pre-trained subject representation for controllable text-to-image generation and editing.
\newblock {\em Advances in Neural Information Processing Systems}, 36:30146--30166, 2023.

\bibitem{liao2024text}
Jiayi Liao, Xu~Chen, Qiang Fu, Lun Du, Xiangnan He, Xiang Wang, Shi Han, and Dongmei Zhang.
\newblock Text-to-image generation for abstract concepts.
\newblock In {\em Proceedings of the AAAI Conference on Artificial Intelligence}, volume~38, pages 3360--3368, 2024.

\bibitem{liu2025llm4gen}
Mushui Liu, Yuhang Ma, Zhen Yang, Jun Dan, Yunlong Yu, Zeng Zhao, Zhipeng Hu, Bai Liu, and Changjie Fan.
\newblock Llm4gen: Leveraging semantic representation of llms for text-to-image generation.
\newblock In {\em Proceedings of the AAAI Conference on Artificial Intelligence}, volume~39, pages 5523--5531, 2025.

\bibitem{liu2023name}
Zhi-Song Liu, Li-Wen Wang, Wan-Chi Siu, and Vicky Kalogeiton.
\newblock Name your style: text-guided artistic style transfer.
\newblock In {\em Proceedings of the IEEE/CVF Conference on Computer Vision and Pattern Recognition}, pages 3530--3534, 2023.

\bibitem{liu2023cones}
Zhiheng Liu, Ruili Feng, Kai Zhu, Yifei Zhang, Kecheng Zheng, Yu~Liu, Deli Zhao, Jingren Zhou, and Yang Cao.
\newblock Cones: Concept neurons in diffusion models for customized generation.
\newblock {\em arXiv preprint arXiv:2303.05125}, 2023.

\bibitem{miao2024subject}
Yanting Miao, William Loh, Suraj Kothawade, Pascal Poupart, Abdullah Rashwan, and Yeqing Li.
\newblock Subject-driven text-to-image generation via preference-based reinforcement learning.
\newblock {\em Advances in Neural Information Processing Systems}, 37:123563--123591, 2024.

\bibitem{mo2024freecontrol}
Sicheng Mo, Fangzhou Mu, Kuan~Heng Lin, Yanli Liu, Bochen Guan, Yin Li, and Bolei Zhou.
\newblock Freecontrol: Training-free spatial control of any text-to-image diffusion model with any condition.
\newblock In {\em Proceedings of the IEEE/CVF Conference on Computer Vision and Pattern Recognition}, pages 7465--7475, 2024.

\bibitem{openai2024gpt4o}
OpenAI.
\newblock Gpt-4o system card.
\newblock \url{https://arxiv.org/abs/2410.21276}, 2024.
\newblock Accessed: 2025-05-10.

\bibitem{peng2024dreambench++}
Yuang Peng, Yuxin Cui, Haomiao Tang, Zekun Qi, Runpei Dong, Jing Bai, Chunrui Han, Zheng Ge, Xiangyu Zhang, and Shu-Tao Xia.
\newblock Dreambench++: A human-aligned benchmark for personalized image generation.
\newblock {\em arXiv preprint arXiv:2406.16855}, 2024.

\bibitem{qu2023layoutllm}
Leigang Qu, Shengqiong Wu, Hao Fei, Liqiang Nie, and Tat-Seng Chua.
\newblock Layoutllm-t2i: Eliciting layout guidance from llm for text-to-image generation.
\newblock In {\em Proceedings of the 31st ACM International Conference on Multimedia}, pages 643--654, 2023.

\bibitem{radford2021learning}
Alec Radford, Jong~Wook Kim, Chris Hallacy, Aditya Ramesh, Gabriel Goh, Sandhini Agarwal, Girish Sastry, Amanda Askell, Pamela Mishkin, Jack Clark, et~al.
\newblock Learning transferable visual models from natural language supervision.
\newblock In {\em International conference on machine learning}, pages 8748--8763. PmLR, 2021.

\bibitem{raffel2020exploring}
Colin Raffel, Noam Shazeer, Adam Roberts, Katherine Lee, Sharan Narang, Michael Matena, Yanqi Zhou, Wei Li, and Peter~J Liu.
\newblock Exploring the limits of transfer learning with a unified text-to-text transformer.
\newblock {\em Journal of machine learning research}, 21(140):1--67, 2020.

\bibitem{raj2023dreambooth3d}
Amit Raj, Srinivas Kaza, Ben Poole, Michael Niemeyer, Nataniel Ruiz, Ben Mildenhall, Shiran Zada, Kfir Aberman, Michael Rubinstein, Jonathan Barron, et~al.
\newblock Dreambooth3d: Subject-driven text-to-3d generation.
\newblock In {\em Proceedings of the IEEE/CVF international conference on computer vision}, pages 2349--2359, 2023.

\bibitem{rout2025semantic}
L~Rout, Y~Chen, N~Ruiz, C~Caramanis, S~Shakkottai, and W~Chu.
\newblock Semantic image inversion and editing using rectified stochastic differential equations.
\newblock In {\em The Thirteenth International Conference on Learning Representations}, 2025.

\bibitem{rout2025rbmodulation}
L~Rout, Y~Chen, N~Ruiz, A~Kumar, C~Caramanis, S~Shakkottai, and W~Chu.
\newblock Rb-modulation: Training-free stylization using reference-based modulation.
\newblock In {\em The Thirteenth International Conference on Learning Representations}, 2025.

\bibitem{ruiz2023dreambooth}
Nataniel Ruiz, Yuanzhen Li, Varun Jampani, Yael Pritch, Michael Rubinstein, and Kfir Aberman.
\newblock Dreambooth: Fine tuning text-to-image diffusion models for subject-driven generation.
\newblock In {\em Proceedings of the IEEE/CVF conference on computer vision and pattern recognition}, pages 22500--22510, 2023.

\bibitem{ryu2023low}
Simo Ryu.
\newblock Low-rank adaptation for fast text-to-image diffusion fine-tuning.
\newblock {\em Low-rank adaptation for fast text-to-image diffusion fine-tuning}, 3, 2023.

\bibitem{sohn2023styledrop}
Kihyuk Sohn, Lu~Jiang, Jarred Barber, Kimin Lee, Nataniel Ruiz, Dilip Krishnan, Huiwen Chang, Yuanzhen Li, Irfan Essa, Michael Rubinstein, et~al.
\newblock Styledrop: Text-to-image synthesis of any style.
\newblock {\em Advances in Neural Information Processing Systems}, 36:66860--66889, 2023.

\bibitem{sun2024autoregressive}
Peize Sun, Yi~Jiang, Shoufa Chen, Shilong Zhang, Bingyue Peng, Ping Luo, and Zehuan Yuan.
\newblock Autoregressive model beats diffusion: Llama for scalable image generation.
\newblock {\em arXiv preprint arXiv:2406.06525}, 2024.

\bibitem{sun2023principle}
Zhiqing Sun, Yikang Shen, Qinhong Zhou, Hongxin Zhang, Zhenfang Chen, David Cox, Yiming Yang, and Chuang Gan.
\newblock Principle-driven self-alignment of language models from scratch with minimal human supervision.
\newblock {\em Advances in Neural Information Processing Systems}, 36:2511--2565, 2023.

\bibitem{voynov2023p+}
Andrey Voynov, Qinghao Chu, Daniel Cohen-Or, and Kfir Aberman.
\newblock p+: Extended textual conditioning in text-to-image generation.
\newblock {\em arXiv preprint arXiv:2303.09522}, 2023.

\bibitem{wang2024instantstyle}
Haofan Wang, Peng Xing, Renyuan Huang, Hao Ai, Qixun Wang, and Xu~Bai.
\newblock Instantstyle-plus: Style transfer with content-preserving in text-to-image generation.
\newblock {\em arXiv preprint arXiv:2407.00788}, 2024.

\bibitem{wang2024taming}
Jiangshan Wang, Junfu Pu, Zhongang Qi, Jiayi Guo, Yue Ma, Nisha Huang, Yuxin Chen, Xiu Li, and Ying Shan.
\newblock Taming rectified flow for inversion and editing.
\newblock {\em arXiv preprint arXiv:2411.04746}, 2024.

\bibitem{wang2024instantid}
Qixun Wang, Xu~Bai, Haofan Wang, Zekui Qin, Anthony Chen, Huaxia Li, Xu~Tang, and Yao Hu.
\newblock Instantid: Zero-shot identity-preserving generation in seconds.
\newblock {\em arXiv preprint arXiv:2401.07519}, 2024.

\bibitem{wang2024genartist}
Zhenyu Wang, Aoxue Li, Zhenguo Li, and Xihui Liu.
\newblock Genartist: Multimodal llm as an agent for unified image generation and editing.
\newblock {\em Advances in Neural Information Processing Systems}, 37:128374--128395, 2024.

\bibitem{wang2023stylediffusion}
Zhizhong Wang, Lei Zhao, and Wei Xing.
\newblock Stylediffusion: Controllable disentangled style transfer via diffusion models.
\newblock In {\em Proceedings of the IEEE/CVF International Conference on Computer Vision}, pages 7677--7689, 2023.

\bibitem{wu2024next}
Shengqiong Wu, Hao Fei, Leigang Qu, Wei Ji, and Tat-Seng Chua.
\newblock Next-gpt: Any-to-any multimodal llm.
\newblock In {\em Forty-first International Conference on Machine Learning}, 2024.

\bibitem{wu2024self}
Tsung-Han Wu, Long Lian, Joseph~E Gonzalez, Boyi Li, and Trevor Darrell.
\newblock Self-correcting llm-controlled diffusion models.
\newblock In {\em Proceedings of the IEEE/CVF Conference on Computer Vision and Pattern Recognition}, pages 6327--6336, 2024.

\bibitem{xai2025grok}
xAI.
\newblock Grok 3 beta — the age of reasoning agents.
\newblock \url{https://x.ai/news/grok-3}, 2025.
\newblock Accessed: 2025-05-10.

\bibitem{xing2024csgo}
Peng Xing, Haofan Wang, Yanpeng Sun, Qixun Wang, Xu~Bai, Hao Ai, Renyuan Huang, and Zechao Li.
\newblock Csgo: Content-style composition in text-to-image generation.
\newblock {\em arXiv preprint arXiv:2408.16766}, 2024.

\bibitem{ye2023ip}
Hu~Ye, Jun Zhang, Sibo Liu, Xiao Han, and Wei Yang.
\newblock Ip-adapter: Text compatible image prompt adapter for text-to-image diffusion models.
\newblock {\em arXiv preprint arXiv:2308.06721}, 2023.

\bibitem{zhang2022dino}
Hao Zhang, Feng Li, Shilong Liu, Lei Zhang, Hang Su, Jun Zhu, Lionel~M Ni, and Heung-Yeung Shum.
\newblock Dino: Detr with improved denoising anchor boxes for end-to-end object detection.
\newblock {\em arXiv preprint arXiv:2203.03605}, 2022.

\bibitem{zhang2023adding}
Lvmin Zhang, Anyi Rao, and Maneesh Agrawala.
\newblock Adding conditional control to text-to-image diffusion models.
\newblock In {\em Proceedings of the IEEE/CVF international conference on computer vision}, pages 3836--3847, 2023.

\bibitem{zhang2024ssr}
Yuxuan Zhang, Yiren Song, Jiaming Liu, Rui Wang, Jinpeng Yu, Hao Tang, Huaxia Li, Xu~Tang, Yao Hu, Han Pan, et~al.
\newblock Ssr-encoder: Encoding selective subject representation for subject-driven generation.
\newblock In {\em Proceedings of the IEEE/CVF Conference on Computer Vision and Pattern Recognition}, pages 8069--8078, 2024.

\bibitem{zhao2023uni}
Shihao Zhao, Dongdong Chen, Yen-Chun Chen, Jianmin Bao, Shaozhe Hao, Lu~Yuan, and Kwan-Yee~K Wong.
\newblock Uni-controlnet: All-in-one control to text-to-image diffusion models.
\newblock {\em Advances in Neural Information Processing Systems}, 36:11127--11150, 2023.

\bibitem{zhu2025internvl3}
Jinguo Zhu, Weiyun Wang, Zhe Chen, Zhaoyang Liu, Shenglong Ye, Lixin Gu, Yuchen Duan, Hao Tian, Weijie Su, Jie Shao, et~al.
\newblock Internvl3: Exploring advanced training and test-time recipes for open-source multimodal models.
\newblock {\em arXiv preprint arXiv:2504.10479}, 2025.

\bibitem{zhu2017unpaired}
Jun-Yan Zhu, Taesung Park, Phillip Isola, and Alexei~A Efros.
\newblock Unpaired image-to-image translation using cycle-consistent adversarial networks.
\newblock In {\em Proceedings of the IEEE international conference on computer vision}, pages 2223--2232, 2017.

\end{thebibliography}
\bibliographystyle{plain}

\newpage
\appendix
\section*{Appendix}
\section{More Baseline Comparisons}
\begin{figure}[h]
    \centering
    \includegraphics[width=0.98\linewidth]{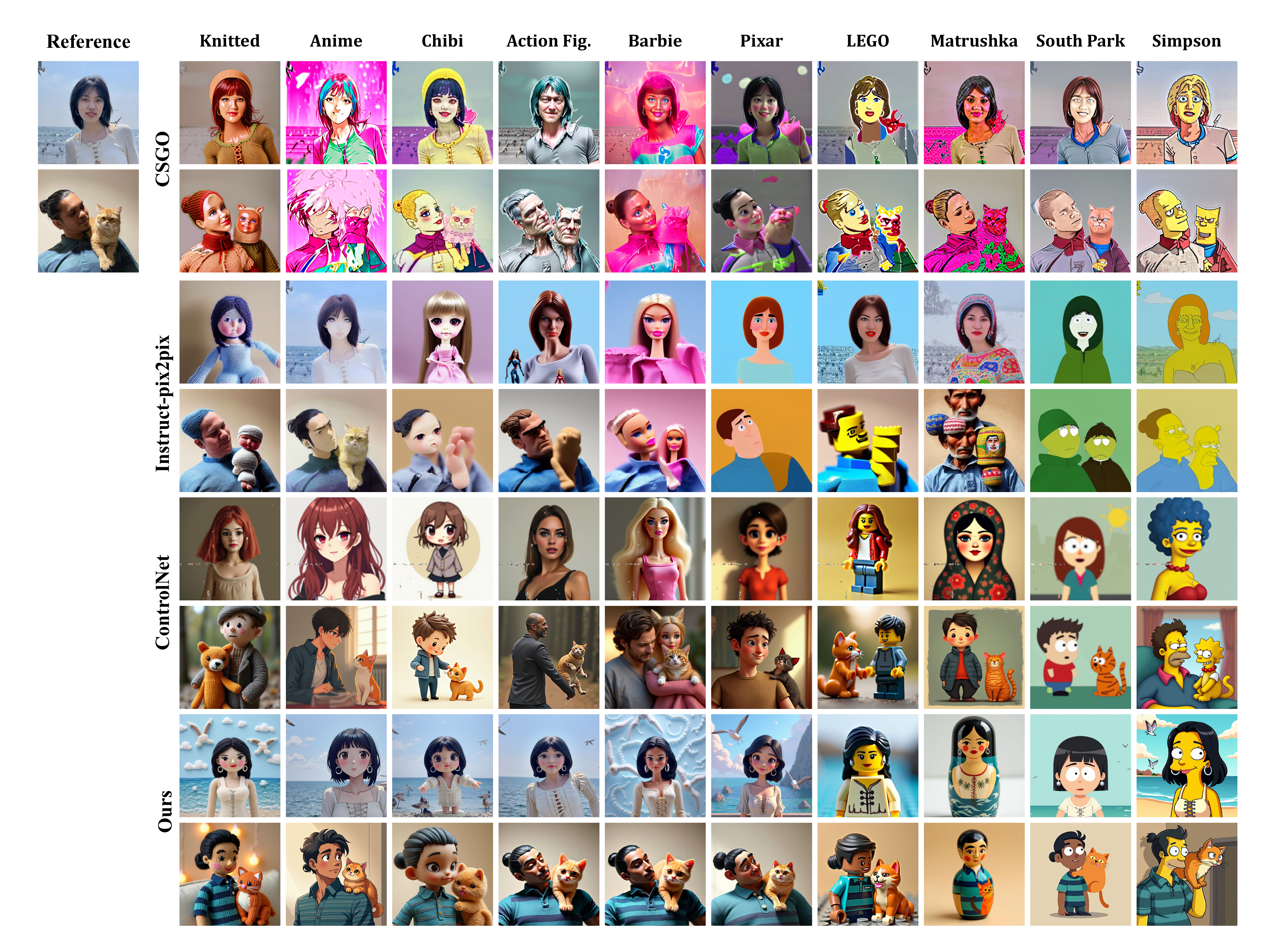}
    \caption{Qualitative comparison with other state-of-the-art methods.}
    \label{fig:baselines}
\end{figure}
We have compared our method with other recent methods and shown the results in Figure~\ref{fig:baselines}


\section{Baselines Setup}
\noindent \textbf{RF-Inversion \cite{rout2024semantic}.} For RF-Inversion experiments, we use the FluxPipeline with pretrained weights from FLUX.1-dev, applying a fixed set of 10 stylized prompts (e.g., Knitted Doll, Lego, Anime, Pixar, Barbie Doll, etc.). Each input image is first inverted into the latent space using 28 inversion steps with a gamma value of 0.5. Stylized outputs are then generated from the inverted latents using the corresponding prompt, with a diffusion range restricted to the first 25\% of the process (start timestep=0.0, stop timestep=0.25) across 50 inference steps. We use eta=0.9 for sampling noise and fix the random seed for deterministic results.

\noindent \textbf{RB-Modulation \cite{wang2024taming}.} We evaluate the RB-Modulation framework, which is built on top of StableCascade, for multi-style portrait generation. The system employs a two-stage cascade: Stage-C integrates style guidance using a reference image via the RBM module, while Stage-B decodes the modulated representation into a high-resolution image. For each input image, we condition generation on a fixed set of ten styles, including \textit{Knitted Doll}, \textit{Lego}, \textit{Anime}, \textit{Pixar}, \textit{South Park}, \textit{Barbie Doll}, among others. Style is injected by computing EffNet embeddings from a style reference image, which are passed to Stage-C. Sampling is performed using a cosine schedule with 20 denoising steps for Stage-C ($\text{cfg}=4.0$) and 10 steps for Stage-B ($\text{cfg}=1.1$). The generation prompt follows the format: \texttt{``a portrait of a [man/woman] in [style] style''}. All images are generated at a resolution of $1024 \times 1024$ and saved under structured directories organized by style. Model weights are loaded from preconfigured files (\texttt{stage\_c\_3b.yaml} and \texttt{stage\_b\_3b.yaml}), and all experiments are run on a single NVIDIA GPU with \texttt{bfloat16} precision.

\noindent \textbf{StyleID \cite{le2022styleid}.} The model performs DDIM inversion of both content and style images to extract intermediate attention features across multiple timesteps and layers. These features are selectively fused by preserving the \texttt{query} components from the content and injecting \texttt{key/value} components from the style via cross-layer modulation. Images are center-cropped and resized to $512 \times 512$, then encoded into a latent space using a 50-step DDIM inversion process. Feature fusion occurs from step $49$ to $0$ using layers \texttt{6} through \texttt{11} of the U-Net architecture. Inference is performed using a modified DDIM sampler with temperature scaling ($T=1.5$) and query preservation strength ($\gamma=0.75$), allowing detailed control over style-content disentanglement. 

\noindent \textbf{InstantID \cite{wang2024instantid}.} We employ the InstantID pipeline, a face-aware stylization system built on top of Stable Diffusion XL with ControlNet and IP-Adapter conditioning. For each image, we detect and extract facial features using the InsightFace `antelopev2` model, which provides both facial landmarks and a 512-dimensional identity embedding. The largest detected face is used for identity preservation. The model is composed of a base SDXL backbone (\texttt{stabilityai/stable-diffusion-xl-base-1.0}), augmented with ControlNet and an IP-Adapter module. Facial keypoints are rendered as a control image, while the identity embedding is injected via the IP-Adapter. The prompt used follows the format: \texttt{``a portrait of [man/woman] in [style] style''}, with gender determined based on pre-mapped image indices. Images are resized to a maximum of $1280 \times 1280$ with alignment to multiples of 64 for efficiency. We apply 30 inference steps with a guidance scale of 5.0, and both the ControlNet and IP-Adapter conditioning scales are set to 0.8. For each image and style (e.g., \textit{Knitted Doll}, \textit{Lego}, \textit{Pixar}, etc.). All experiments are performed on a CUDA-enabled GPU with float32 precision.

\noindent \textbf{DreamBooth \cite{ruiz2023dreambooth}.} We fine-tune a Stable Diffusion model using the DreamBooth method for subject-driven generation. The model is trained with a unique identifier token associated with a specific subject. During inference, we use the fine-tuned model to generate stylized outputs conditioned on prompts such as \texttt{``a photo of sks person in south park style''}. We utilize the \texttt{StableDiffusionPipeline} from the \texttt{diffusers} library with half-precision (float16) enabled for memory efficiency. Classifier-free guidance is set to 7.5, and inference is performed with 50 sampling steps. Each prompt is repeated four times to generate a batch of diverse samples. All experiments are conducted on an NVIDIA GPU with CUDA acceleration.

\noindent \textbf{CSGO \cite{xing2024csgo}.} The model is built on top of Stable Diffusion XL and integrates IP-Adapter modules for content and style injection, in conjunction with ControlNet for structural guidance. We use a curated set of exemplar images representing ten artistic styles (\textit{Knitted Doll}, \textit{Lego}, \textit{Anime}, \textit{Pixar}, \textit{Barbie Doll}, etc.). For each input image, we determine a prompt of the form \texttt{``a portrait of [man/woman] in [style]''} based on predefined subject-gender mappings. For multi-subject images, we use a generalized prompt: \texttt{``convert into [style] style''}. The pipeline uses pretrained SDXL weights, a customized VAE, and a ControlNet variant trained on tiling patterns. Identity conditioning is injected via content and style tokens (\texttt{4} and \texttt{32} tokens respectively), targeting both base UNet and ControlNet blocks. The content image is resized to align with the model’s base resolution and passed through a shared resampler. Stylization is conditioned on a ControlNet scale of 0.6, with guidance and style scales set to 10.0 and 1.0, respectively. Generation is performed over 50 denoising steps, and the seed is fixed to 42 for deterministic outputs.

\noindent \textbf{DiffArtist \cite{jiang2024diffartist}.} DiffArtist is built on Stable Diffusion and SDXL. We use 50 DDIM inversion and generation steps with a classifier-free guidance scale of 3.5. Attention injection is configured with \texttt{share\_key=True}, \texttt{share\_query=True}, \texttt{share\_value=False}, \texttt{share\_resnet\_layers=[0,1]}, \texttt{share\_attn=False}, and \texttt{share\_cross\_attn=False}. AdaIN is enabled (\texttt{use\_adain=True}), and content anchoring is applied during disentangled generation (\texttt{use\_content\_anchor=True}). SDXL runs in \texttt{float16} precision, while the VAE operates in \texttt{float32}.

\section{Prompt for Identity Distillation}

\begin{tcolorbox}[colback=gray!10, colframe=gray!50, boxrule=0.5mm, rounded corners, title={\textbf{\textcolor{black}{Facial Attributes.}}}]
I will provide you with an image of a human face. Produce a comprehensive, forensic style facial description in a single dense paragraph, detailed enough for accurate human reconstruction or forensic sketching. Describe the individual’s apparent biological sex, race or ethnic background, estimated age range, and general physique or body weight appearance (e.g., underweight, average, heavyset, muscular). Detail the skin tone with precision note undertones (cool, warm, neutral), pigmentation, and visible texture such as freckles, moles, scars, or blemishes. For the face shape, specify whether it is oval, round, square, heartshaped, or diamond, and describe the jawline in terms of sharpness, width, and angularity. Include the forehead’s height and width, the hairline contour, and whether it is straight, receding, or widow’s peak. The cheekbones should be described with respect to height, prominence, and lateral placement. Indicate the chin’s form pointed, cleft, square, rounded, recessed, or protruding. For the nose, describe the overall size, bridge contour (e.g., high, flat, concave, convex), nostril flare, tip shape, and symmetry. Then describe the eyes in terms of shape (almond, round, hooded, monolid), tilt (upturned, downturned, straight), size, spacing, iris color, scleral clarity, and the presence or absence of an eyelid crease. Note eyelash length and curl if prominent. Detail the eyebrows’ shape (arched, flat, curved), thickness, length, color, grooming style, and position relative to the eye socket. For the mouth, describe lip shape (e.g., heartshaped, bowshaped), fullness, vertical distance to nose and chin, symmetry, the sharpness or softness of the cupid’s bow, and overall coloring or pigmentation. Describe the teeth if visible alignment, spacing, color. Include a description of the ears, noting size, angle of protrusion, lobe attachment, and visibility from the front. Describe hair color, texture (straight, wavy, curly, coiled), length, parting, volume, and any dyeing or graying. Finally, include the inter feature ratios, such as the distance from eyes to nose, nose to mouth, mouth to chin, and face width to height, noting vertical and horizontal harmony or asymmetry. Use anatomically precise, observational language as would be found in professional forensic profiling.
\end{tcolorbox}

\begin{tcolorbox}[colback=gray!10, colframe=gray!50, boxrule=0.5mm, rounded corners, title={\textbf{\textcolor{black}{Clothes and Accessories.}}}]
I will provide you with an image of a human subject. Generate a highly detailed, forensic style description of the subject’s clothing, accessories, and appearance related adornments in a single continuous paragraph. The description should allow an observer to accurately reconstruct the subject’s outfit, materials, and stylistic presentation. Begin with the overall outfit type, including the layering of garments and whether the attire suggests a formal, casual, professional, athletic, cultural, or weather specific context. Describe each clothing item from top to bottom identify the garment type (e.g., jacket, blouse, hoodie, dress), cut and silhouette (e.g., fitted, oversized, cropped, tailored), and the fabric type (cotton, denim, silk, wool, synthetic blend) including texture (smooth, ribbed, coarse, sheer, glossy, matte). Include color, noting primary tones, secondary hues, patterns (e.g., floral, plaid, geometric, logo prints), and fading or distressing. Specify visible closures such as buttons, zippers, drawstrings, or snaps, and describe collars, sleeves, cuffs, hems, stitching, and trim in detail. Mention the fit and drape on the body, whether loose, structured, or body hugging. Describe pants, skirts, or lower garments similarly, including waistband type, length, cut (e.g., tapered, flared, pleated), and material behavior (e.g., stiff, flowy, elastic). Provide precise detail on footwear, covering type (e.g., sneakers, boots, sandals), color, condition (new, worn, scuffed), sole thickness, fasteners, and branding if visible. For accessories, describe all items such as belts, hats, scarves, bags, jewelry, watches, or sunglasses, including their material (leather, metal, plastic), placement, size, color, and design. For jewelry, note shape, stone types, metal color, and position (e.g., left wrist, right earlobe). Mention any logos, emblems, name tags, or inscriptions, their location, style, and legibility. Include hair style related adornments like hair clips, ties, bands, or veils if present. Indicate whether the clothing is clean, wrinkled, tailored, or weathered, and how it interacts with the subject’s posture or movement. Use clear, descriptive language with observational precision, suitable for law enforcement or forensic documentation.
\end{tcolorbox}

\begin{tcolorbox}[colback=gray!10, colframe=gray!50, boxrule=0.5mm, rounded corners, title={\textbf{\textcolor{black}{Posture.}}}]
I will provide you with an image of a human subject. Generate an anatomically precise, forensic style description of the subject’s pose, body orientation, and posture in a single dense paragraph, suitable for reconstruction in forensic modeling, animation, or figure drawing. Begin by describing the overall stance whether the subject is standing, sitting, leaning, crouching, walking, or in motion and specify the distribution of body weight (e.g., evenly balanced, shifted to one hip, resting on one leg, slouched). Indicate the torso orientation, such as facing forward, three quarter turn, side profile, or twisted at the waist. Describe the spine posture in terms of straightness, curvature, or slouch. Detail shoulder position (level, raised, dropped, angled) and arm placement, specifying whether arms are relaxed, crossed, bent, akimbo, behind the back, or holding an object. Note the hand position, including whether fingers are spread, curled, pointing, or interacting with any surface, accessory, or garment. Include the leg position and angle whether straight, crossed, bent, staggered, or one knee slightly raised and describe foot placement relative to the body and ground, noting if feet are parallel, angled outward/inward, or midstride. Specify the head tilt, pitch, yaw, and roll (e.g., upright, nodding forward, tilted to the side, turned partially), and describe eye gaze direction (e.g., looking straight ahead, offcamera, downward, or toward an object or person). Indicate whether the pose appears relaxed, tense, alert, casual, formal, or dynamic, and if the subject is interacting with the environment (e.g., leaning against a wall, sitting on a chair, walking up stairs). Include relational geometry such as angles between limbs, distance between hands and torso, and chin to shoulder alignment to allow for 3D pose reconstruction. Use clear anatomical and kinesiological language appropriate for forensic modeling or biomechanical analysis.
\end{tcolorbox}

\begin{tcolorbox}[colback=gray!10, colframe=gray!50, boxrule=0.5mm, rounded corners, title={\textbf{\textcolor{black}{Background.}}}]
I will provide you with an image containing a human subject within a visible environment. Generate a comprehensive, continuous paragraph that describes the background and setting in exhaustive detail, suitable for recreating the scene in forensic reconstruction, virtual staging, or cinematic layout. Begin with the type of environment  whether it is indoor or outdoor, public or private, residential, commercial, natural, or constructed. Describe the primary spatial context such as a street, room, park, studio, hallway, or landscape, and specify depth cues like perspective, vanishing points, and visible horizon lines. Detail the lighting conditions, indicating whether the light is natural or artificial, the light source direction (e.g., overhead, frontal, backlit), shadow presence and length, and overall ambience (e.g., bright, dim, moody, clinical). Provide the color palette of the background  dominant hues, gradients, saturation, and color temperature (warm, cool, neutral). Describe surface textures and materials  walls, floors, or ground surfaces (e.g., tiled, carpeted, wooden, concrete, grassy), their cleanliness, reflectiveness, or damage. Include any visible structures or objects, such as furniture, windows, fences, signage, artwork, cables, curtains, vehicles, trees, or architectural details, describing their placement, size, condition, and interaction with light or shadows. Mention background activity if any  other people, movement, animals, or vehicles  and whether the environment is static or dynamic. Indicate the depth of field  how blurred or sharp background elements appear relative to the subject. Include information about weather or atmospheric effects (e.g., fog, rain, sunlight, haze, reflections), time of day, and seasonal indicators (e.g., dry leaves, snow, blossoming plants). Describe any visible text, logos, posters, or signage in the background, noting legibility, language, and style. Conclude with the spatial relationship between the subject and background  how far the subject appears from walls, objects, or vanishing points  and whether there are any visual elements framing or isolating the subject. Use precise, observational language suitable for forensic, cinematic, or architectural analysis.
\end{tcolorbox}

\begin{tcolorbox}[colback=gray!10, colframe=gray!50, boxrule=0.5mm, rounded corners, title={\textbf{\textcolor{black}{Combine.}}}]
I will provide a detailed description of an image. Your task is to compress and distill the description into two versions:

\noindent \textbf{T5 Embedding Prompt (512 tokens max):}
 Use all 512 tokens. Preserve the most salient, semantically rich, and distinctive features of the image that would help T5 encode identity, context, style, and layout. Discard irrelevant or negative information (e.g., "no visible tattoos") if necessary for brevity.

\noindent \textbf{CLIP Embedding Prompt (77 tokens max):}
 Create a compressed caption and use all 77 tokens, optimized for CLIP's contrastive text-image embedding. Prioritize visual discriminability and identity-relevant details, especially those that can anchor visual features (e.g., colors, shapes, facial attributes, clothing style, environment cues).

In both cases, your goal is to retain the most distinguishing and generative elements from the input description, suitable for retrieval or conditional generation tasks.
\end{tcolorbox}

\section{StyleBench Evaluation Prompt}

\begin{tcolorbox}[colback=gray!10, colframe=gray!50, boxrule=0.5mm, rounded corners, title={\textbf{\textcolor{black}{Prompt.}}}]

\noindent \textbf{Task Definition}
You will be provided with three inputs: a reference image, a stylized generated image, and a style prompt. Your task is to evaluate how well the generated image captures the intended style abstraction while preserving the recognizable identity of the subject in the reference image.

\noindent \textbf{Evaluation Focus}
This is a task of stylized abstraction, not realism or direct replication. The goal is not to retain exact facial proportions or textures from the reference but to abstract the subject into a distinct artistic or toy-like style. The reference image provides the identity cues, such as hairstyle, accessories, clothing, skin tone, or posture, which should be recognizable in the generated image despite heavy stylistic transformation. The style prompt dictates the visual language that the generation should adhere to, such as the yarn texture of a knitted doll, the plastic blockiness of a LEGO figure, or the yellow skin and cartoon geometry of a Simpson character.
A good abstraction interprets the identity through the lens of the target style. For instance, in a LEGO doll style, the facial features may become minimal and geometric, but an iconic element like Einstein’s wild hair or Michael Jackson’s fedora must still be present. In a South Park style, the flat shading and round cut-out forms are expected, while the individual identity is retained through clothing or hair color. Likewise, in Ghibli or Van Gogh styles, the abstraction involves painterly texture and expressive strokes rather than exact geometry. The fidelity of abstraction lies in this balance between stylization and recognizability.

\noindent \textbf{Scoring Criteria}
You must assess the generated image based on three integrated dimensions:
\begin{itemize}
\item How well the image adheres to the specified style in the prompt.

\item Whether key identifying features from the reference image are preserved.

\item Whether the abstraction effectively fuses identity and style into a coherent result.
\end{itemize}

Note that expressions, pose, and compositional cues should come from the reference image. The quality of the abstraction depends on both a strong stylistic transformation and the faithful reinterpretation of subject identity.

\noindent \textbf{Scoring Range}
Assign a score from 0 to 4:

Very Poor (0): The generated image fails to apply the intended style and does not retain recognizable identity.

Poor (1): The style is weak or incorrect; most identity traits are lost.
Fair (2): The style is moderately applied, with partial identity preservation.
Good (3): The image reflects the correct style and retains most identity features.
Excellent (4): The image is a faithful and expressive stylization that captures both the identity and style seamlessly.

\noindent \textbf{Input Format}
You will receive:

A reference image

A stylized generated image

A style prompt

\noindent \textbf{Output Format}

\noindent \textbf{Score: [Your Score]}

Only return the score. Do not include any justification or explanation.

\end{tcolorbox}

\end{document}